\documentclass[letterpaper]{article} 
\usepackage{aaai2026}  
\usepackage{times}  
\usepackage{helvet}  
\usepackage{courier}  
\usepackage[hyphens]{url}  
\usepackage{graphicx} 
\usepackage{natbib}  
\usepackage{caption} 
\usepackage{algorithm}
\usepackage{algorithmic}

\usepackage[utf8]{inputenc}
\usepackage{amsfonts}
\usepackage{nicefrac}
\usepackage{amsmath}
\usepackage{booktabs}
\usepackage{threeparttable}
\usepackage{array}
\usepackage{subcaption}
\usepackage{xcolor}        
\usepackage{tcolorbox}
\usepackage{comment}
\usepackage{todonotes}

\usepackage{newfloat}
\usepackage{listings}
\DeclareCaptionStyle{ruled}{labelfont=normalfont,labelsep=colon,strut=off} 
\floatstyle{ruled}
\newfloat{listing}{tb}{lst}{}
\floatname{listing}{Listing}
\title{LLMs for Game Theory: Entropy-Guided In-Context Learning and Adaptive CoT Reasoning}

\author {
    Tommaso Felice Banfi\textsuperscript{\rm 1},
    Sashenka Gamage\textsuperscript{\rm 2}
}

\affiliations {
    \textsuperscript{\rm 1}Politecnico di Milano, Dept. of Electronics, Information and Bioengineering, Milan, Italy.\\
    \textsuperscript{\rm 2}The Hong Kong Polytechnic University, Dept. of Electrical and Electronics Engineering, Hong Kong, China.\\
    tommasofelice.banfi@mail.polimi.it, sashenka.gamage@connect.polyu.hk
}

\begin{document}

\maketitle

\begin{abstract}
	We propose a novel LLM-based framework for reasoning in discrete, game-theoretic tasks, illustrated with \emph{Tic-Tac-Toe}.
	The method integrates in-context learning with entropy-guided chain-of-thought (CoT) reasoning and adaptive context retrieval.
	The model dynamically adjusts both the number of retrieved examples and reasoning paths according to token-level uncertainty:
	concise reasoning with minimal context is used when uncertainty is low, whereas higher
	uncertainty triggers expanded multi-path CoT exploration.
	Experimental evaluation against a sub-optimal algorithmic opponent shows that entropy-aware adaptive reasoning
	substantially improves decision quality, increasing the average game outcome from \(-11.6\%\) with the baseline LLM
	to \(+9.5\%\) with entropy-guided adaptive reasoning over 100 games (win = +1, tie = 0, loss = -1),
	while maintaining a relatively low number of LLM queries per game.
	Statistical validation confirms that the improvement is significant, and correlation analysis reveals
	a negative association between token-level entropy and move optimality.
	These findings demonstrate that uncertainty-guided adaptive reasoning effectively enhances LLM performance
	in sequential decision-making environments.
\end{abstract}


\section{Introduction}
Large Language Models (LLMs) have recently shown impressive capabilities across a wide range of reasoning and knowledge-intensive tasks.
They perform well in single-step inference, language understanding, and few-shot generalization.
However, they still face notable challenges in structured, sequential decision-making problems, where each move affects all future states and outcomes.
In such settings, effective reasoning requires not only local consistency but also long-term planning and adaptive use of prior knowledge.

In this work, we explore how to improve the performance of LLMs in sequential reasoning environments by combining two complementary mechanisms.
First, we present \emph{entropy-guided context retrieval}, a method that dynamically adjusts the
number of retrieved examples based on the model's predictive uncertainty.
When the model shows low entropy (high confidence), it works with a compact and focused context;
when uncertainty increases, it automatically retrieves semantically similar past states and
their corresponding optimal actions, extending its in-context knowledge.
Second, we introduce an \emph{adaptive chain-of-thought (CoT)} reasoning framework, where the
model uses concise, single-path reasoning when confident and branches into multiple reasoning paths when token-level entropy indicates uncertainty.
Together, these mechanisms enable efficient, uncertainty-aware reasoning, focusing computational effort only when necessary.

We use \emph{Tic-Tac-Toe} as a controlled and fully observable testbed to evaluate our approach.
Despite its simplicity, the game captures key aspects of sequential decision-making (state transitions,
opponent modeling, and outcome optimization) while allowing objective evaluation of move quality through the minimax algorithm.
This setup supports a rigorous and quantitative assessment of LLM reasoning and provides statistical validation of the proposed methods.

The rest of this paper is organized as follows.
Section~\ref{sec:related_work} reviews related research on reasoning with LLMs, in-context learning, and uncertainty estimation.
Section~\ref{sec:problem_formulation} defines the game-theoretic setting and evaluation framework.
Our proposed approach, including entropy-guided context retrieval and adaptive CoT reasoning, is presented in Section~\ref{sec:method}.
Section~\ref{sec:experimental_setup_and_results} describes the experimental setup, baselines, and metrics,
followed by a detailed analysis of the results in Section~\ref{sec:results_and_discussion}.
Section~\ref{sec:limitations_and_future_work} outlines current limitations and potential directions for future research.
Finally, Section~\ref{sec:conclusion} concludes the paper, highlighting the broader impact of uncertainty-aware
reasoning for LLMs in sequential decision-making tasks.

\section{Related Work}
\label{sec:related_work}

Prior works in the NLP research community has shown that chain-of-thought prompting improves
LLM performance on complex reasoning tasks by exposing intermediate reasoning
steps~\cite{wei2023chainofthoughtpromptingelicitsreasoning}.
In addition, frameworks such as Tree-of-Thoughts implement branching
over “thought” states (i.e., multi-path reasoning~\cite{zhou2024enhancing}
and tree-based CoT~\cite{bi2024forest}) to perform deliberative search with
LLMs, demonstrating substantial performance gains in non-trivial
tasks~\cite{yao2023treethoughtsdeliberateproblem}.

However, evaluating CoT and related reasoning frameworks remains an open challenge.
Most of the methods focus on getting the answer, rather than evaluating the quality of the reasoning
method. Benchmarking studies ~\cite{golovneva2022roscoe,prasad2023receval,jacovi2024chainofthought}
have shown that CoT reasoning performance can vary widely across tasks and domains, and that
standardized evaluation metrics are often insufficient to capture nuanced reasoning quality.
A recent survey~\cite{lee2025evaluatingstepby} emphasize that assessing CoT effectiveness in
open-ended reasoning tasks is inconsistent due to the fragmented progress over evaluation design
and benchmark development. In our setting, this limitation is mitigated, as each game state in
\emph{Tic-Tac-Toe} has a well-defined optimal move due to the underlying game theory.
This allows us to quantitatively evaluate step-by-step reasoning quality by comparing model-generated moves
against the algorithmic optimal strategy obtained algorithmically.

In line with that, retrieval-augmented generation
methods (RAG)~\cite{lewis2021retrievalaugmentedgenerationknowledgeintensivenlp} have shown
that giving examples at inference time improves knowledge-intensive tasks and reduces
the burden on the model's internal parameters.
Finally, uncertainty-aware or entropy-guided reasoning
approaches~\cite{yao2023treethoughtsdeliberateproblem, wang2023selfconsistencyimproveschainthought, zhang2025entropy, jia2025entropy},
adaptively decide when to expand the LLM reasoning depth or breadth (based on the model's uncertainty).
Recent work also highlights the synergy between RAG and in-context learning (ICL), where the strong
inductive feature bias of ICL can be reduced by RAG, injecting external information at inference
time, providing more relevant examples
or facts~\cite{shi2023rethinkingicl}.
As we discussed before, evaluating only final answers in multi-step reasoning often
obscures the nuanced reasoning errors that models make.
For instance, GridPuzzle~\cite{tyagi2024stepbystepreasoningsolvegrid} develops a dataset and evaluation
framework to analyze reasoning chains in grid-based puzzles, identifying where models like
GPT-4, Claude-3, and LLaMA-2 falter.
Similarly, the EIC (Error Identification and Correction) benchmark~\cite{li2024evaluatingmathematicalreasoninglarge} evaluates mathematical reasoning
by highlighting error types and enabling models to correct mistakes with targeted prompts.
These works emphasize the importance of truly measuring reasoning
ability beyond overall accuracy.

Other methods, such as RCOT (Reversing Chain-of-Thought)~\cite{xue2023rcot} and
Verify-and-Edit~\cite{zhao2023verifyandedit}, focus on test-time correction by detecting
inconsistencies in the generated reasoning chains and guiding LLMs to revise them.
RCOT performs a backtracking approach by reconstructing problems from solutions to
find factual inconsistencies, while Verify-and-Edit introduces knowledge-aware editing
to improve CoT reliability.

Classical game AI research has long relied on search algorithms such as
minimax~\cite{russell2020aima} and alpha-beta pruning~\cite{knuth1975analysis}.
With the advent of deep learning, systems like AlphaZero have shown that integrating learned policy networks
with planning can outperform pure search-based engines in complex games like
chess and Go~\cite{silver2017masteringchessshogiselfplay}. The Alympics (“Olympics for Agents”), a
simulation framework that uses LLM agents to study game-theoretic interactions, offers a methodology
for constructing strategic games, defining evaluation criteria~\cite{mao2025alympics}.
It highlights issues of strategic reasoning, adaptation, and how LLMs can simulate human-like strategic interactions.
By incorporating entropy-based measures of uncertainty~\cite{wang2023selfconsistencyimproveschainthought, ltsz23inform},
one can identify decisions where LLM agents are less confident and adaptively expand reasoning paths,
reduce hallucination, and improve efficiency by focusing computation where uncertainty is high.
Building on these insights, our approach integrates retrieval-augmented prompting with entropy-guided
chain-of-thought reasoning to simulate agentic decision-making in fully observable game environments,
enabling both rigorous evaluation and controlled sub-optimality testing.

In our case, the game of \emph{Tic-Tac-Toe} is simple enough that a full minimax solution is feasible and
provides the optimal moves for all possible board states.
Our work lies at the intersection, we combined a learned vector representation
of game states with retrieval-augmented prompting and an entropy-guided chain-of-thought strategy.

\section{Problem Definition}
\label{sec:problem_formulation}

We consider discrete, two-player, turn-based games, a standard framework in game theory for sequential decision-making under perfect information. A game is defined by the tuple $(S, A, T, R)$, where:
\begin{itemize}
	\item $S$ denotes the set of possible game states (board configurations),
	\item $A(s)$ is the set of legal actions available in state $s \in S$,
	\item $T: S \times A \rightarrow S$ represents the deterministic transition function, and
	\item $R: S \times A \rightarrow \mathbb{R}$ is the reward function.
\end{itemize}

We focus on \emph{Tic-Tac-Toe}, a finite, zero-sum, perfect-information game played on a $3 \times 3$ grid.
Players alternate placing their marker (X or O) on an empty cell. A player wins by completing a horizontal,
vertical, or diagonal line of three markers, while a draw occurs if all cells are filled without a winning configuration.

Formally, the state space is $S = \{s_1, s_2, \dots, s_{3^9}\}$, and for each state $s$, the action space
is defined as $A(s) = \{(i, j) \mid \text{cell } (i, j) \text{ is empty}\}$. The game evolves according to
the transition function $s_{t+1} = T(s_t, a_t)$ at each turn $t$.

In our setting, the LLM acts as one of the players, competing against a fixed suboptimal algorithmic opponent.
Its performance is evaluated based on the distribution of outcomes (win, tie, and loss) across multiple games.
During play, the model can retrieve contextual information from a limited database of board states and
corresponding optimal actions to inform its decisions.

\section{Method}
\label{sec:method}
In this section, we present the proposed framework for LLM-based decision-making in \emph{Tic-Tac-Toe}.
The approach integrates two complementary components: \emph{entropy-guided context retrieval} and
\emph{adaptive chain-of-thought reasoning}, designed to improve efficiency and robustness in sequential reasoning tasks.

The overall architecture follows a game loop where the LLM observes the current board state, generates
a reasoning trace, and selects the next action based on both its internal confidence and retrieved
contextual information. Although developed and evaluated in the context of turn-based games, the
framework is general and can be extended to broader sequential decision-making scenarios involving LLMs.

\subsection{Game Loop}
\label{sec:game_loop}

The match is initialized with an empty board $B$, where the LLM plays as ``X'' and the algorithmic agent as ``O''.
The starting player is selected randomly, and the game proceeds as follows:

\begin{enumerate}
	\item The algorithmic agent selects a move according to its policy and updates the board $B$.
	\item A structured prompt is constructed for the LLM, including the current board configuration,
	      available moves, the identity of the active player, and a set of retrieved examples with their
	      corresponding optimal actions (see Section~\ref{sec:board_representation_and_context_retrieval}).
	\item The prompt is passed to $\text{LLM}(x_q, p, \mathcal{R}_q)$, and the model’s textual output is obtained.
	\item The output is parsed to extract move coordinates. If the move is invalid, the LLM is prompted to regenerate.
	      If it remains invalid, a random valid move is selected.
	\item If a terminal state (win, loss, or tie) is reached, the game ends; otherwise, the loop continues from step~1.
\end{enumerate}

\subsection{Board Representation and Context Retrieval}
\label{sec:board_representation_and_context_retrieval}

Following Tyagi et al.~\cite{tyagi2024stepbystepreasoningsolvegrid}, we represent the \emph{Tic-Tac-Toe} board as a $3 \times 3$ tensor
\begin{equation}
	B \in \{0,1,2\}^{3 \times 3},
\end{equation}
where $0$ denotes an empty cell, $1$ represents the marker ``X'', and $2$ represents ``O''. To facilitate LLM-based retrieval, we flatten $B$ into a vector
\begin{equation}
	x = \text{vec}(B) \in \mathbb{R}^9
\end{equation}
and encode it into a lower-dimensional latent representation $z \in \mathbb{R}^d$ using an autoencoder:
\begin{align}
	z       & = f_\theta(x), \\
	\hat{x} & = g_\phi(z),
\end{align}
where $f_\theta: \mathbb{R}^9 \to \mathbb{R}^d$ is the encoder and $g_\phi: \mathbb{R}^d \to \mathbb{R}^9$ is the decoder.
The reconstruction loss ensures that the latent space preserves the essential information of the board:
\begin{equation}
	\mathcal{L}_{\text{rec}}(\theta, \phi) = \| x - g_\phi(f_\theta(x)) \|^2.
\end{equation}

To enable effective context retrieval, we further structure the latent space using a contrastive learning objective.
Given a pair of boards $(x_i, x_j)$ with associated optimal moves $(m_i, m_j)$, the contrastive loss encourages latent vectors of boards
with the same optimal move to be close, while pushing apart boards with different optimal moves:
\begin{equation}
	\mathcal{L}_{\text{con}} =
	\begin{cases}
		\| f_\theta(x_i) - f_\theta(x_j) \|^2,                           & \text{if } m_i = m_j,    \\
		\max \big(0, \tau - \| f_\theta(x_i) - f_\theta(x_j) \| \big)^2, & \text{if } m_i \neq m_j,
	\end{cases}
\end{equation}
where $\tau > 0$ is a margin hyperparameter controlling the minimum distance between latent representations of boards with different optimal moves.
This formulation ensures that similar board states are grouped in the latent space according to the optimal action, enabling efficient retrieval of
relevant past states for in-context reasoning.

The final training objective combines the reconstruction and contrastive components:
\begin{equation}
	\mathcal{L} = \mathcal{L}_{\text{rec}} + \lambda \mathcal{L}_{\text{con}},
\end{equation}
where $\lambda > 0$ balances the contribution of the contrastive loss relative to the reconstruction loss.
Minimizing $\mathcal{L}$ results in a latent space that simultaneously preserves board structure and organizes states according to optimal moves,
supporting uncertainty-aware context retrieval for LLM reasoning.

\paragraph{Vector Database.}
A vector database is constructed as
\begin{equation}
	\mathcal{D} = \{(z_i, m_i)\},
\end{equation}
where each entry consists of a latent embedding $z_i = f_\theta(x_i)$ paired with its corresponding optimal move $m_i$.
In practice, approximately $20\%$ of all possible board states are stored in $\mathcal{D}$, selected uniformly at random,
together with their minimax-optimal moves.

At inference time, given a query board $x_q$, we first compute its embedding
\begin{equation}
	z_q = f_\theta(x_q).
\end{equation}
The retrieval set $\mathcal{R}_q$ is then constructed by selecting the $k$ most similar examples from $\mathcal{D}$ according to cosine similarity:
\begin{align}
	\text{sim}(z_q, z_i) & = \frac{z_q \cdot z_i}{\|z_q\| \, \|z_i\|},                                                     \\
	\mathcal{R}_q        & = \text{Top-}k \{ (x_i, m_i) \in \mathcal{D} \;\big|\; \text{sim}(z_q, z_i) \text{ largest} \}.
\end{align}

To adapt retrieval to the model's uncertainty, we employ an \emph{entropy-aware} mechanism detailed in
Section~\ref{sec:adaptive_chain_of_thought_reasoning}.

The retrieval size $k$ is dynamically adjusted according to
\begin{equation}
	k = \min \big( k_{\max}, \; \lceil k_0 + \alpha \cdot H_q \rceil \big),
\end{equation}
where $k_0$ is the base number of examples, $\alpha > 0$ is a scaling factor, and $k_{\max}$ is a hard cap determined by the maximum available
context window of the LLM.
Consequently, high-entropy outputs trigger the retrieval of additional examples to enrich the context, while low-entropy outputs limit the
retrieval to a small set of nearest neighbors, preserving context space for reasoning.
In all cases, the size of the final LLM context $\mathcal{C}_q$ is bounded by the token budget
$|\mathcal{C}_q| \leq L_{\max}$,
where $L_{\max}$ denotes the maximum number of tokens the model can condition on.

\paragraph{Context Construction.}
The final context $\mathcal{C}_q$ provided to the LLM consists of three components:
\begin{enumerate}
	\item The current board configuration $x_q$,
	\item The identity of the active player $p \in \{\text{X}, \text{O}\}$, and
	\item The retrieved set $\mathcal{R}_q = \{(x_{i_j}, m_{i_j})\}_{j=1}^k$, containing the $k$ most similar boards and their
	      corresponding optimal moves.
\end{enumerate}
These elements are concatenated and formatted into a structured prompt, which is then passed as input to the LLM, see Appendix~\ref{sec:example_prompt} for an example of the prompt.
This structured context ensures that the model can leverage both the current board state and relevant past
experiences to make informed decisions.

\subsection{Adaptive Chain-of-Thought Reasoning}
\label{sec:adaptive_chain_of_thought_reasoning}

In our framework, \emph{chain-of-thought} (CoT) reasoning is employed to model sequential decision-making in \emph{Tic-Tac-Toe}.
Each CoT step corresponds to a game turn, and intermediate reasoning involves simulating both the current player's move and the
opponent's response. Formally, let $s_t \in S$ denote the board state at turn $t$, and let the CoT generate a sequence
of state-action pairs:
\begin{equation}
	\text{CoT} = \big( (s_0, a_0), (s_1, a_1), \dots, (s_T, a_T) \big),
\end{equation}
where $a_t \in A(s_t)$ is the move selected by the active player or the opponent at step $t$, and $T$ is the horizon
until a terminal state is reached.

\paragraph{Reasoning Modes.}
We define several CoT strategies with increasing complexity:

\begin{itemize}
	\item \textbf{Direct output}: the LLM outputs a single move $a_t$ for the current state $s_t$ without simulating future steps.
	      Mathematically, the selected action is
	      \begin{equation}
		      a_t = \arg\max_{a \in A(s_t)} P_{\text{LLM}}(a \mid s_t, \mathcal{C}_t),
	      \end{equation}
	      where $\mathcal{C}_t$ is the context provided to the LLM.

	\item \textbf{Multi-CoT}: the LLM generates $n$ independent single-path sequences $\text{CoT}^{(j)}$ from the current
	      state until a terminal state is reached.
	      The final move is selected by a majority vote over the first actions $a_0^{(j)}$ of each sequence:
	      \begin{equation}
		      a_t = \text{mode} \{ a_0^{(j)} \}_{j=1}^n.
	      \end{equation}

	\item \textbf{Tree-based CoT}: at each step, the LLM generates $n$ candidate actions for the active player, and for
	      each candidate, all possible responses of the opponent are simulated, forming a tree of depth $d$:
	      \begin{equation}
		      \mathcal{T}_t = \{ (s_{t+1}^{(i)}, a_{t+1}^{(i)}) \}_{i=1}^{n \cdot |A(s_{t+1})|}.
	      \end{equation}
	      The evaluation function $V: S \to \mathbb{R}$ (entropy based selection) is used to select the
	      top-$k$ branches at each level:
	      \begin{equation}
		      \mathcal{B}_t = \text{Top-}k \{ \mathcal{T}_t \mid V(s_{t+1}^{(i)}) \text{ is maximal} \}.
	      \end{equation}

	\item \textbf{Entropy-guided CoT}: a hybrid approach in which multiple reasoning branches are generated only when
	      the LLM exhibits high uncertainty. Uncertainty is quantified by the token-level entropy of the CoT output (see below).
	      Formally, for a reasoning step $t$, if $H_t^{\text{step}} > H_{\text{th}}$, the model generates $n_t$ parallel branches:
	      \begin{equation}
		      n_t = \min \big( n, |A(s_t)| \big),
	      \end{equation}
	      where $|A(s_t)|$ is the number of legal actions at $s_t$.
\end{itemize}

\paragraph{Token-level Entropy and Adaptive Branching.}
At each reasoning step $t$, the LLM generates a sequence of tokens $\{v_{t,k}\}_{k=1}^{L_t}$, each associated with a
probability distribution over the vocabulary $\{p_{t,k}^{(i)}\}_{i=1}^{|V|}$. The token-level entropy is computed as
\begin{equation}
	\label{eq:token_entropy}
	H_{t,k}^{\text{token}} = - \sum_{i=1}^{|V|} p_{t,k}^{(i)} \log p_{t,k}^{(i)},
\end{equation}
and the average token entropy defines the step-level entropy:
\begin{equation}
	\label{eq:step_entropy}
	H_t^{\text{step}} = \frac{1}{L_t} \sum_{k=1}^{L_t} H_{t,k}^{\text{token}}.
\end{equation}

To adaptively determine the number of reasoning branches, we define a set of ordered entropy thresholds
\begin{equation}
	0 = H_0 < H_1 < \dots < H_m,
\end{equation}
where $m$ is the number of threshold levels. Each threshold $H_j$ corresponds to a predefined number of branches $n_j$, forming a mapping
\begin{equation}
	H_t^{\text{step}} \in [H_j, H_{j+1}) \;\; \Rightarrow \;\; n_t = n_j, \quad j = 0, \dots, m-1.
\end{equation}
In this way, higher step-level entropy triggers the generation of more reasoning branches, while low entropy results in fewer
branches, optimizing computational resources.

The actual number of branches is selected as
\begin{equation}
	n_t = \min\big(n_j, |A(s_t)|\big),
\end{equation}
where $|A(s_t)|$ is the number of legal actions at state $s_t$.

Once generated, each branch represents a candidate CoT trajectory that is recursively expanded. To control the computational
cost, only the top-$k$ branches are retained at each step according to a scoring function $S(\text{CoT})$
based on the average entropy of the tokens in the reasoning path:
\begin{equation}
	\mathcal{B}_t = \text{Top-}k \{ \text{CoT branches at step } t \mid S(\text{CoT}) \}.
\end{equation}

This adaptive threshold mechanism allows the model to selectively explore multiple reasoning paths only when the uncertainty
is high, providing a principled balance between reasoning quality and computational efficiency. Future extensions may consider
dynamically updating thresholds $H_j$ based on the distribution of entropies observed across previous steps, rather than
using fixed values.

\section{Experimental Setup}
\label{sec:experimental_setup_and_results}

We evaluate the proposed LLM agent against a suboptimal algorithmic opponent in \emph{Tic-Tac-Toe}.
Each game continues until a win or a tie. The opponent employs a precomputed Minimax table to rank
all legal moves according to their optimality.

\begin{table*}[!ht]
	\centering
	\caption{Performance of the LLM agent against an algorithmic opponent using the LLaMA-7B model.
		Each cell shows the average outcome per game $S$ over 100 games
		and the rounded average number of LLM queries
		per game. Positive $S$ values indicate that the LLM agent won
		more games than it lost (e.g., +10 means 10\% more wins than losses), and
		vice versa for negative values.}

	\label{tab:results}
	\begin{tabular}{l*{6}{>{\raggedleft\arraybackslash}p{1.2cm}}}
		\toprule
		                            & \multicolumn{2}{>{\raggedleft\arraybackslash}p{2.4cm}}{No Additional Context}
		                            & \multicolumn{2}{>{\raggedleft\arraybackslash}p{2.4cm}}{Fixed-Size Context}
		                            & \multicolumn{2}{>{\raggedleft\arraybackslash}p{2.5cm}}{\textbf{Entropy-Guided Context}}                                                              \\
		                            & $S$ [\%]                                                                                & Queries & $S$ [\%] & Queries & $S$ [\%]      & Queries     \\
		\midrule
		No CoT                      & -11.6                                                                                   & 3       & -5.2     & 4       & -2.8          & 4           \\

		Single CoT                  & -8.2                                                                                    & 13      & -2.6     & 13      & -0.1          & 15          \\

		Multi CoT                   & -7.5                                                                                    & 24      & -1.2     & 26      & +4.8          & 28          \\

		Tree-based CoT              & -2.7                                                                                    & 165     & +4.5     & 178     & +9.8          & 188         \\

		\textbf{Entropy-Guided CoT} & -4.1                                                                                    & 48      & +3.8     & 56      & \textbf{+9.5} & \textbf{48} \\

		\bottomrule
	\end{tabular}
\end{table*}

Let $n$ denote the number of legal moves, and let $r_i \in \{1, \dots, n\}$ be the rank of move $i$
(with $r_i = 1$ corresponding to the most optimal move, and $r_i = n$ to the least optimal).
Given a skill level $\alpha \in [0,1]$ (set to $\alpha = 0.95$ in our experiments), the probability
of selecting move $i$ is defined as
\begin{equation}
	P(i) = \frac{\max\big(0, 1 - |r_i - \alpha n|/(n-1)\big)}{\sum_{j=1}^{n} \max\big(0, 1 - |r_j - \alpha n|/(n-1)\big)}.
\end{equation}
This distribution peaks at the move corresponding to the agent's skill level and decays linearly
toward both less optimal and overly optimal moves, ensuring that the agent plays moves consistent
with its skill on average while introducing controlled randomness.
Since \emph{Tic-Tac-Toe} is a solved game, statistical validation of the strategy is unnecessary,
as the Minimax table provides a complete ranking of valid moves for any board state.

The LLM employs parallel and tree-based chain-of-thought (CoT) reasoning. In our experiments, we
configured $n=3$ parallel CoT sequences, each allowing up to $3$ branches per turn, and retain at
most $k=10$ top paths at each step to control computational cost.

All experiments utilize the LLaMA-7B model~\cite{touvron2023llamaopenefficientfoundation, touvron2023llama2openfoundation}
to establish baseline performance.
This model was chosen to balance inference capability and computational resource requirements, as further discussed
in Section~\ref{sec:limitations_and_future_work}.

We report two primary metrics per configuration:
\begin{enumerate}
	\item The average game outcome $S$ over 100 games, expressed as a percentage, with scores assigned as win = +1,
	      tie = 0, and loss = -1.
	\item The average number of LLM queries per game, including additional calls required for multi-path reasoning,
	      recomputation due to invalid outputs, and context expansion under high uncertainty.
\end{enumerate}

For the first move of the game, due to board symmetry and multiple equally optimal moves, entropy-based branching
is not applied and all equally optimal moves are considered valid without adjustment.

\begin{figure*}[!ht]
	\centering
	\includegraphics[width=0.70\textwidth]{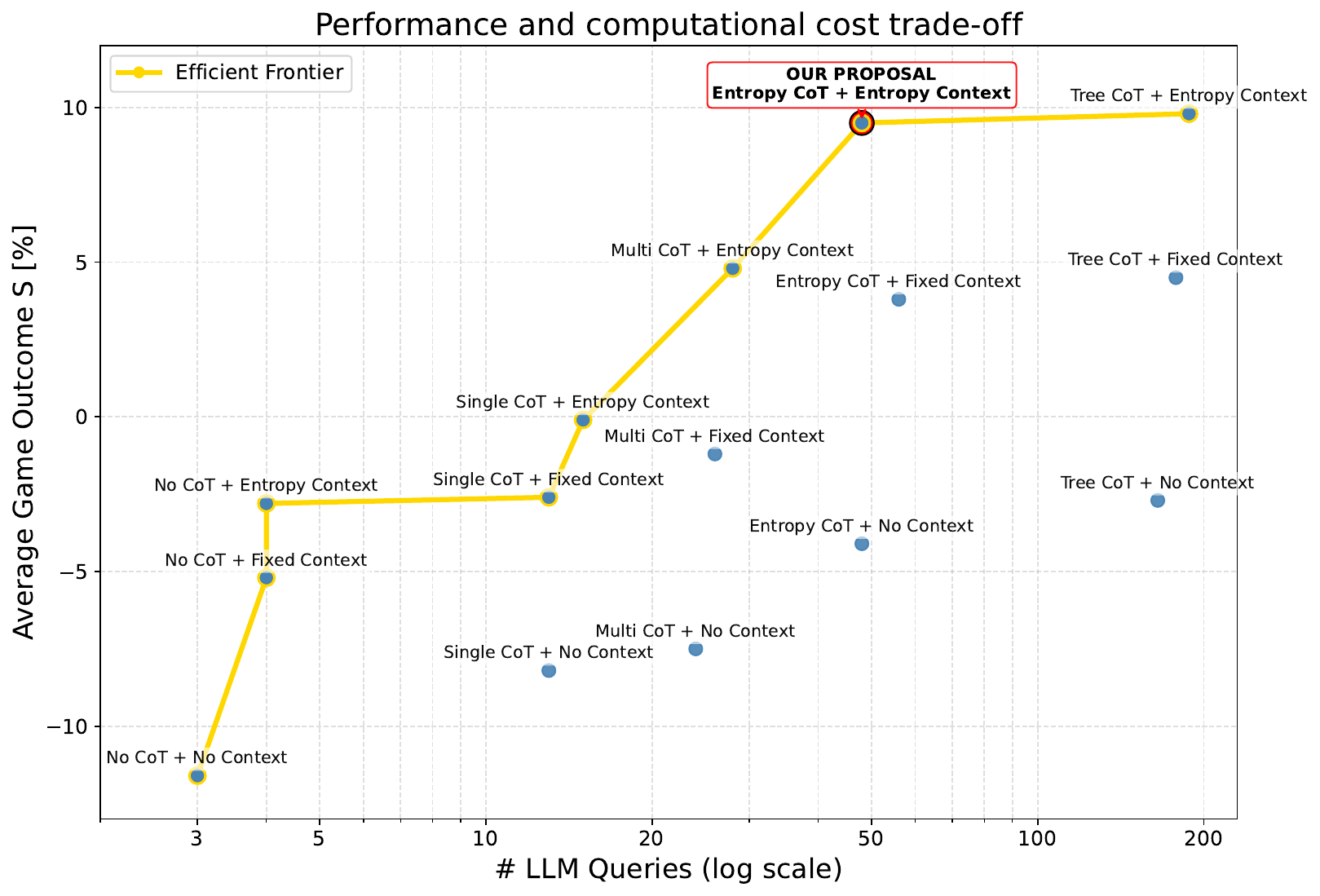}
	\caption{Comparison of retrieval-based in-context learning methods with different CoT strategies, evaluated in terms of average game outcome $S$ and computational cost (number of LLM queries). The plot highlights the performance--cost trade-off and the resulting efficient frontier across methods.}
	\label{fig:performance_cost_trade-off}
\end{figure*}

\paragraph{Hyperparameters and Implementation Details.}
All experiments employed the LLaMA-7B model~\cite{touvron2023llamaopenefficientfoundation} to establish
baseline performance.
For initial testing, we used the smaller Gemma 3 270M model~\cite{gemmateam2025gemma3technicalreport} on
a single NVIDIA GeForce RTX 4050 with 6~GB VRAM and an Intel i7-12700H CPU with 16~GB RAM.
Final evaluations with LLaMA-7B~\cite{touvron2023llamaopenefficientfoundation} were
conducted on an NVIDIA GeForce RTX 4060 Ti with 16~GB VRAM, sufficient
for inference-only tasks. The operating system used was Ubuntu 24.04 LTS.

For next-move generation, LLM hyperparameters were set to ensure deterministic and precise outputs:
temperature = 0.1, top-p and top-k sampling disabled (top\_p=1.0, top\_k=0), and beam search set to 2.
The maximum number of generated tokens per move was 10, covering all possible coordinate formats.
Padding and end-of-sequence tokens were assigned to the tokenizer's \texttt{eos\_token\_id}.

A fixed random seed of 42 was used to ensure reproducibility in selecting and ordering retrieval examples
(see Section~\ref{sec:board_representation_and_context_retrieval}).
Each combination of context type and CoT reasoning (3 context types $\times$ 5 CoT strategies = 15 configurations)
was initially evaluated using the Gemma 3 model.
Final evaluations employed LLaMA-7B with 100 games per configuration against the algorithmic opponent, measuring
both average game outcome $S$ and the number of LLM queries per game. Hyperparameters were selected based on
preliminary experiments to guarantee consistent generation of valid move coordinates.

The implementation was developed in Python using the \texttt{transformers} library for model and tokenizer
management, \texttt{bitsandbytes} for quantization, \texttt{torch} as the main framework, and \texttt{langchain}
to handle context retrieval and CoT reasoning.

\section{Results and Discussion}
\label{sec:results_and_discussion}
Our results in Table~\ref{tab:results} and Figure~\ref{fig:performance_cost_trade-off} show that both context retrieval and chain-of-thought (CoT) reasoning
substantially affect the LLM agent's performance. Without additional context, the agent performs poorly, with
negative average outcomes across all CoT strategies. Introducing a fixed-size context improves performance for
all strategies by providing relevant examples that guide move selection. Entropy-guided context retrieval further
enhances outcomes by dynamically adjusting the number of retrieved examples based on predictive uncertainty,
allowing the model to access more relevant information when needed. For example, using entropy-guided context
without CoT improves the outcome from $-11.6\%$ to $-2.8\%$. Multi CoT generates multiple candidate reasoning
paths, and tree-based CoT explores a broader set of future states and opponent responses, achieving the highest
performance (+9.8\%) but with a substantially higher number of queries (188 on average). Combining entropy-guided
CoT with entropy-aware context retrieval achieves nearly equivalent performance (+9.5\%) while requiring only
48 queries, roughly one-fourth of the tree-based CoT cost. These results demonstrate that adaptive context
retrieval and selective multi-path reasoning enable efficient, high-quality decision-making with significantly
reduced computational overhead.

\paragraph{Entropy as a Proxy for Uncertainty.}

Token-level entropy has been recently used as a measure of model uncertainty in discrete multi-step reasoning
tasks~\cite{zhang2025entropy, jia2025entropy, ltsz23inform}.
It quantifies the uncertainty of the predicted probability distribution over the next token: low entropy
indicates a concentrated distribution on a few tokens, while high entropy corresponds to a more uniform
distribution over multiple plausible tokens.

\begin{figure*}[!ht]
	\centering
	\includegraphics[width=0.66\textwidth]{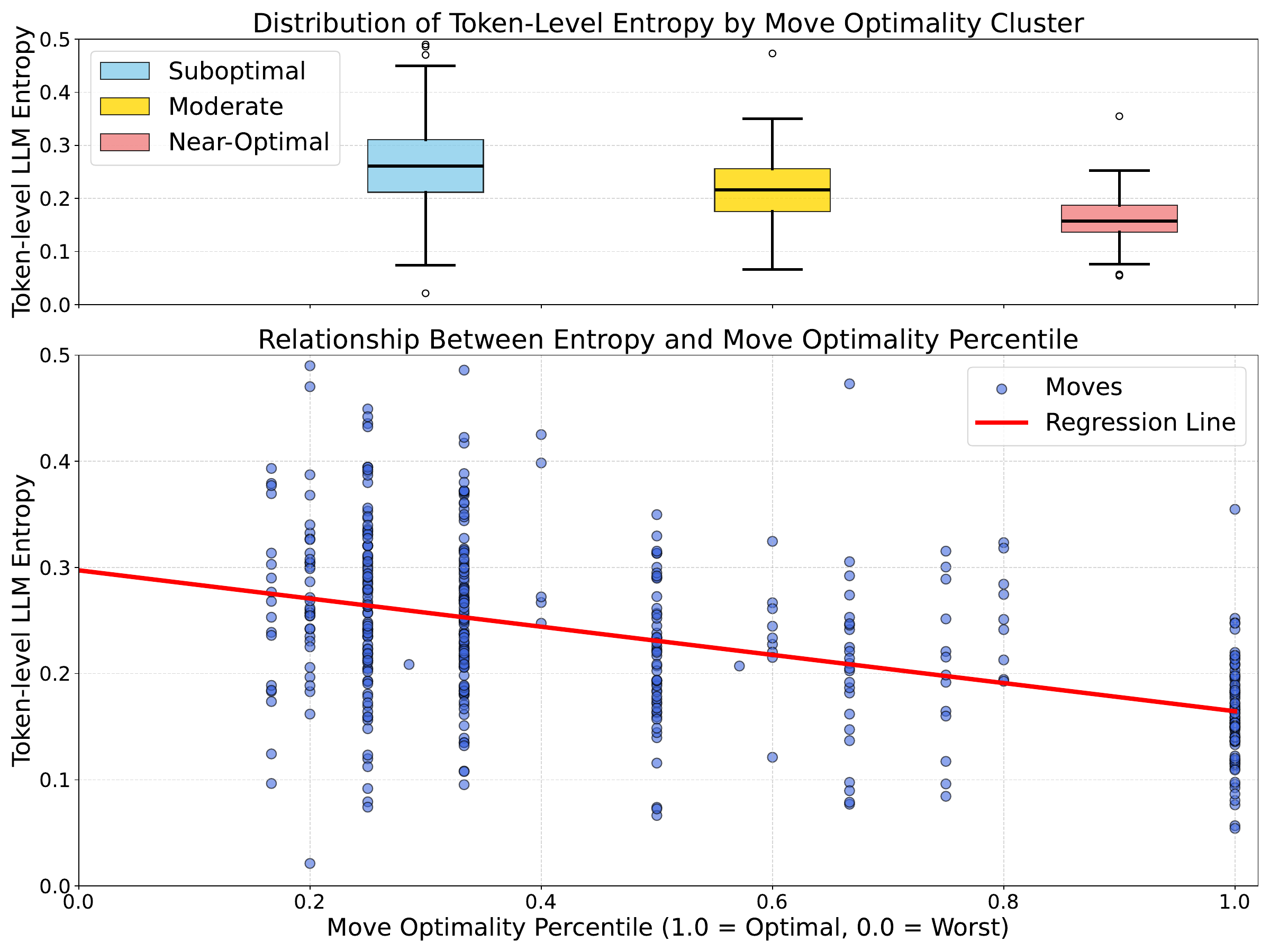}
	\caption{Token-level entropy versus move optimality percentile. Top: boxplots grouped into three
		clusters (\emph{Suboptimal}, \emph{Moderate}, \emph{Near-Optimal}), showing higher entropy for
		less optimal moves. Bottom: scatter plot of individual moves with a linear regression line,
		illustrating the negative relationship between entropy and move optimality.
		Data are based on 500 randomly selected board states.} 
	\label{fig:entropy_vs_optimality}
\end{figure*}

We evaluated the LLM on a random subset of 500 valid board states, generating the next move without CoT or
additional context and recording the token-level entropy of each output, as defined in Equation~\ref{eq:step_entropy}. For each move, we computed its ranking percentile relative to all valid moves in the same state. Figure~\ref{fig:entropy_vs_optimality} shows the relationship between token-level entropy and move optimality percentile.

High entropy arises primarily in two cases: (i) genuine uncertainty about the next move (true positive),
or (ii) multiple moves with similar optimality (false positive). False positives are mostly observed in
early-game states and are mitigated by avoiding entropy-based branching on the first move.
In general, high entropy consistently indicates potentially suboptimal choices.

To quantify this noisy relationship, we computed rank-based correlations between entropy and move optimality percentiles.
Spearman correlation is $\rho = -0.471$, $p < 10^{-3}$, and Kendall Tau correlation
is $\tau = -0.346$, $p < 10^{-3}$, both statistically significant. These results confirm a consistent
negative trend: moves with higher entropy are more likely to be suboptimal.

This analysis supports the use of token-level entropy as a proxy for uncertainty in sequential decision-making tasks.

\paragraph{Discussion on Zero-Shot Claim.}
The results support the hypothesis that retrieval-augmented in-context learning (RA-ICL) enables effective
zero-shot reasoning in structured decision-making tasks such as \emph{Tic-Tac-Toe}.
Although the retrieval encoder described in Section~\ref{sec:board_representation_and_context_retrieval}
was trained in a self-supervised manner on domain data, the objective of learning semantically consistent
representations of board states is task-agnostic and does not involve supervision on optimal actions.
Thus, the LLM is never fine-tuned for the downstream task, and reasoning emerges purely from contextual
conditioning provided by retrieval.

Unlike fine-tuning approaches, which modify model parameters, retrieval-augmented inference operates
externally: relevant examples are dynamically retrieved and injected into the prompt based on latent
similarity. This mechanism aligns with large-scale
architectures such as GPT-5~\cite{openai2025gpt5} and Gemini~2.5~\cite{google2024gemini2_5}, where
retrieval modules enable adaptive context composition 
accessing vast external knowledge bases
(e.g., web documents, code repositories, etc.) without model fine-tuning.

Empirically, the model exhibits context-driven reasoning that scales with the adaptivity of the retrieval
process. The entropy-guided mechanism allows selective context expansion only under high uncertainty,
reducing unnecessary computation while maintaining decision quality.
Overall, the proposed architecture satisfies the key zero-shot conditions: no task-specific fine-tuning,
self-supervised retrieval embedding construction, and dynamic in-context adaptation driven by model
uncertainty.

These results indicate that retrieval-augmented chain-of-thought reasoning constitutes a viable and scalable zero-shot framework for structured problem-solving.

\section{Limitations and Future Work}
\label{sec:limitations_and_future_work}

The proposed framework has several limitations that define directions for future work.

First, all experiments were conducted on \emph{Tic-Tac-Toe}, a deterministic and fully observable game
with a small state and action space. This simplified setup allows controlled evaluation but does not
capture the challenges of larger or stochastic environments. Extending the framework to more complex
games such as Connect Four or Go would test scalability and robustness under higher-dimensional state
representations and longer reasoning sequences.

Second, the model used (LLaMA-7B~\cite{touvron2023llamaopenefficientfoundation}) has limited reasoning
capacity. This helps isolate the effects of adaptive chain-of-thought and context retrieval, but stronger
models (e.g., GPT-5~\cite{openai2025gpt5}, Gemini~2.5~\cite{google2024gemini2_5}) may already handle part
of this reasoning intrinsically. Future work should test whether entropy-guided mechanisms still offer
benefits when the base model is more capable.

Third, the use of token-level entropy assumes a direct link between linguistic uncertainty and decision
uncertainty. While supported by correlation results, this assumption may not hold in settings where
uncertainty arises from ambiguity rather than lack of knowledge. Alternative uncertainty estimators,
such as ensemble variance or mutual information, could provide a more reliable signal.

Fourth, the retrieval database was generated from the full game tree and may not generalize to domains
with incomplete or biased data. Retrieval redundancy or conflict between examples could also affect
the model’s reasoning consistency.

Finally, the evaluation focused on quantitative performance (win/loss and query count) without assessing
logical soundness or interpretability of the reasoning traces. Extending analysis to partially observable
tasks and studying reasoning quality under uncertainty remain open directions.

Future work will address these limitations by scaling the approach to larger games and more general
decision-making tasks, integrating improved uncertainty estimation, and evaluating reasoning behavior
beyond accuracy-based metrics.

\section{Conclusion}
\label{sec:conclusion}

We introduced a framework for sequential decision-making with large language models that combines
entropy-guided context retrieval with adaptive chain-of-thought reasoning. By selectively expanding
reasoning depth and contextual information under uncertainty, the method improves decision quality
while maintaining a low number of LLM queries. In our experiments on \emph{Tic-Tac-Toe}, this approach
increased the average game outcome from -11.6\% to +9.5\%, demonstrating that uncertainty-aware
adaptation effectively enhances model consistency and efficiency.

Although tested on a simple, fully observable game, the framework is model-agnostic and generalizable
to larger or partially observable domains. Its core principle of dynamically allocating reasoning effort
based on uncertainty offers a scalable method for integrating retrieval-augmented reasoning and
adaptive CoT in more complex decision-making or reasoning tasks where zero-shot performance remains
limited.

Future extensions will explore applying this mechanism to domains with non-deterministic
dynamics and incomplete information, where uncertainty-guided reasoning may provide significant
benefits in both performance and interpretability.

\section*{Acknowledgments}
The author would like to thank the anonymous reviewers for their valuable feedback. This work was developed by the author T.F. Banfi while also affiliated with Huawei Milan/Maxwell Research Center, but it is independent and not related to his work there. This research did not receive any specific grant.

\bibliography{aaai2026}

@misc{wei2023chainofthoughtpromptingelicitsreasoning,
  title        = {Chain-of-Thought Prompting Elicits Reasoning in Large Language Models},
  author       = {Jason Wei and Xuezhi Wang and Dale Schuurmans and Maarten Bosma and Brian Ichter and Fei Xia and Ed Chi and Quoc Le and Denny Zhou},
  year         = {2023},
  eprint       = {2201.11903},
  archivePrefix= {arXiv},
  primaryClass = {cs.CL},
  url          = {https://arxiv.org/abs/2201.11903},
}

@misc{yao2023treethoughtsdeliberateproblem,
      title={Tree of Thoughts: Deliberate Problem Solving with Large Language Models}, 
      author={Shunyu Yao and Dian Yu and Jeffrey Zhao and Izhak Shafran and Thomas L. Griffiths and Yuan Cao and Karthik Narasimhan},
      year={2023},
      eprint={2305.10601},
      archivePrefix={arXiv},
      primaryClass={cs.CL},
      url={https://arxiv.org/abs/2305.10601}, 
}

@misc{lewis2021retrievalaugmentedgenerationknowledgeintensivenlp,
      title={Retrieval-Augmented Generation for Knowledge-Intensive NLP Tasks}, 
      author={Patrick Lewis and Ethan Perez and Aleksandra Piktus and Fabio Petroni and Vladimir Karpukhin and Naman Goyal and Heinrich Küttler and Mike Lewis and Wen-tau Yih and Tim Rocktäschel and Sebastian Riedel and Douwe Kiela},
      year={2021},
      eprint={2005.11401},
      archivePrefix={arXiv},
      primaryClass={cs.CL},
      url={https://arxiv.org/abs/2005.11401}, 
}

@misc{silver2017masteringchessshogiselfplay,
      title={Mastering Chess and Shogi by Self-Play with a General Reinforcement Learning Algorithm}, 
      author={David Silver and Thomas Hubert and Julian Schrittwieser and Ioannis Antonoglou and Matthew Lai and Arthur Guez and Marc Lanctot and Laurent Sifre and Dharshan Kumaran and Thore Graepel and Timothy Lillicrap and Karen Simonyan and Demis Hassabis},
      year={2017},
      eprint={1712.01815},
      archivePrefix={arXiv},
      primaryClass={cs.AI},
      url={https://arxiv.org/abs/1712.01815}, 
}

@misc{wang2023selfconsistencyimproveschainthought,
      title={Self-Consistency Improves Chain of Thought Reasoning in Language Models}, 
      author={Xuezhi Wang and Jason Wei and Dale Schuurmans and Quoc Le and Ed Chi and Sharan Narang and Aakanksha Chowdhery and Denny Zhou},
      year={2023},
      eprint={2203.11171},
      archivePrefix={arXiv},
      primaryClass={cs.CL},
      url={https://arxiv.org/abs/2203.11171}, 
}

@misc{gemmateam2025gemma3technicalreport,
      title={Gemma 3 Technical Report}, 
      author={Gemma Team and Aishwarya Kamath and Johan Ferret and Shreya Pathak and Nino Vieillard and Ramona Merhej and Sarah Perrin and Tatiana Matejovicova and Alexandre Ramé and Morgane Rivière and Louis Rouillard and Thomas Mesnard and Geoffrey Cideron and Jean-bastien Grill and Sabela Ramos and Edouard Yvinec and Michelle Casbon and Etienne Pot and Ivo Penchev and Gaël Liu and Francesco Visin and Kathleen Kenealy and Lucas Beyer and Xiaohai Zhai and Anton Tsitsulin and Robert Busa-Fekete and Alex Feng and Noveen Sachdeva and Benjamin Coleman and Yi Gao and Basil Mustafa and Iain Barr and Emilio Parisotto and David Tian and Matan Eyal and Colin Cherry and Jan-Thorsten Peter and Danila Sinopalnikov and Surya Bhupatiraju and Rishabh Agarwal and Mehran Kazemi and Dan Malkin and Ravin Kumar and David Vilar and Idan Brusilovsky and Jiaming Luo and Andreas Steiner and Abe Friesen and Abhanshu Sharma and Abheesht Sharma and Adi Mayrav Gilady and Adrian Goedeckemeyer and Alaa Saade and Alex Feng and Alexander Kolesnikov and Alexei Bendebury and Alvin Abdagic and Amit Vadi and András György and André Susano Pinto and Anil Das and Ankur Bapna and Antoine Miech and Antoine Yang and Antonia Paterson and Ashish Shenoy and Ayan Chakrabarti and Bilal Piot and Bo Wu and Bobak Shahriari and Bryce Petrini and Charlie Chen and Charline Le Lan and Christopher A. Choquette-Choo and CJ Carey and Cormac Brick and Daniel Deutsch and Danielle Eisenbud and Dee Cattle and Derek Cheng and Dimitris Paparas and Divyashree Shivakumar Sreepathihalli and Doug Reid and Dustin Tran and Dustin Zelle and Eric Noland and Erwin Huizenga and Eugene Kharitonov and Frederick Liu and Gagik Amirkhanyan and Glenn Cameron and Hadi Hashemi and Hanna Klimczak-Plucińska and Harman Singh and Harsh Mehta and Harshal Tushar Lehri and Hussein Hazimeh and Ian Ballantyne and Idan Szpektor and Ivan Nardini and Jean Pouget-Abadie and Jetha Chan and Joe Stanton and John Wieting and Jonathan Lai and Jordi Orbay and Joseph Fernandez and Josh Newlan and Ju-yeong Ji and Jyotinder Singh and Kat Black and Kathy Yu and Kevin Hui and Kiran Vodrahalli and Klaus Greff and Linhai Qiu and Marcella Valentine and Marina Coelho and Marvin Ritter and Matt Hoffman and Matthew Watson and Mayank Chaturvedi and Michael Moynihan and Min Ma and Nabila Babar and Natasha Noy and Nathan Byrd and Nick Roy and Nikola Momchev and Nilay Chauhan and Noveen Sachdeva and Oskar Bunyan and Pankil Botarda and Paul Caron and Paul Kishan Rubenstein and Phil Culliton and Philipp Schmid and Pier Giuseppe Sessa and Pingmei Xu and Piotr Stanczyk and Pouya Tafti and Rakesh Shivanna and Renjie Wu and Renke Pan and Reza Rokni and Rob Willoughby and Rohith Vallu and Ryan Mullins and Sammy Jerome and Sara Smoot and Sertan Girgin and Shariq Iqbal and Shashir Reddy and Shruti Sheth and Siim Põder and Sijal Bhatnagar and Sindhu Raghuram Panyam and Sivan Eiger and Susan Zhang and Tianqi Liu and Trevor Yacovone and Tyler Liechty and Uday Kalra and Utku Evci and Vedant Misra and Vincent Roseberry and Vlad Feinberg and Vlad Kolesnikov and Woohyun Han and Woosuk Kwon and Xi Chen and Yinlam Chow and Yuvein Zhu and Zichuan Wei and Zoltan Egyed and Victor Cotruta and Minh Giang and Phoebe Kirk and Anand Rao and Kat Black and Nabila Babar and Jessica Lo and Erica Moreira and Luiz Gustavo Martins and Omar Sanseviero and Lucas Gonzalez and Zach Gleicher and Tris Warkentin and Vahab Mirrokni and Evan Senter and Eli Collins and Joelle Barral and Zoubin Ghahramani and Raia Hadsell and Yossi Matias and D. Sculley and Slav Petrov and Noah Fiedel and Noam Shazeer and Oriol Vinyals and Jeff Dean and Demis Hassabis and Koray Kavukcuoglu and Clement Farabet and Elena Buchatskaya and Jean-Baptiste Alayrac and Rohan Anil and Dmitry and Lepikhin et al. 
      },
      year={2025},
      eprint={2503.19786},
      archivePrefix={arXiv},
      primaryClass={cs.CL},
      url={https://arxiv.org/abs/2503.19786}, 
}

@misc{zhou2024enhancing,
title={Enhancing the General Agent Capabilities of Low-Parameter LLMs through Tuning and Multi-Branch Reasoning}, 
      author={Qinhao Zhou and Zihan Zhang and Xiang Xiang and Ke Wang and Yuchuan Wu and Yongbin Li},
      year={2024},
      eprint={2403.19962},
      archivePrefix={arXiv},
      primaryClass={cs.CL},
      url={https://arxiv.org/abs/2403.19962}, 
}

@misc{bi2024forest,
      title={Forest-of-Thought: Scaling Test-Time Compute for Enhancing LLM Reasoning}, 
      author={Zhenni Bi and Kai Han and Chuanjian Liu and Yehui Tang and Yunhe Wang},
      year={2025},
      eprint={2412.09078},
      archivePrefix={arXiv},
      primaryClass={cs.CL},
      url={https://arxiv.org/abs/2412.09078}, 
}

@misc{zhang2025entropy,
      title={Entropy-based Exploration Conduction for Multi-step Reasoning}, 
      author={Jinghan Zhang and Xiting Wang and Fengran Mo and Yeyang Zhou and Wanfu Gao and Kunpeng Liu},
      year={2025},
      eprint={2503.15848},
      archivePrefix={arXiv},
      primaryClass={cs.AI},
      url={https://arxiv.org/abs/2503.15848}, 
}

@inproceedings{jia2025entropy,
author={Jia, Yongjian and Cai, Jianping and Liu, Ximeng},
  booktitle={2025 3rd International Conference on Big Data and Privacy Computing (BDPC)}, 
  title={Entropy-Guided Tree of Thoughts: A Dynamic Approach to Diverse Path Generation in LLM Reasoning}, 
  year={2025},
  volume={},
  number={},
  pages={132-136},
  doi={10.1109/BDPC63545.2025.11135640}}

@article{knuth1975analysis,
title = {An analysis of alpha-beta pruning},
journal = {Artificial Intelligence},
volume = {6},
number = {4},
pages = {293-326},
year = {1975},
issn = {0004-3702},
doi = {https://doi.org/10.1016/0004-3702(75)90019-3},
url = {https://www.sciencedirect.com/science/article/pii/0004370275900193},
author = {Donald E. Knuth and Ronald W. Moore}
}

@techreport{openai2025gpt5,
  title        = {GPT-5 System Card},
  author       = {{OpenAI}},
  year         = {2025},
  month        = aug,
  institution  = {OpenAI},
  note         = {Version with system card, August 13, 2025},
  url          = {https://cdn.openai.com/gpt-5-system-card.pdf} 
}

@techreport{google2024gemini2_5,
  title        = {Gemini 2.5: Pushing the Frontier with Advanced Reasoning, Multimodality, Long Context, and Next Generation Agentic Capabilities},
  author       = {{Gemini Team, Google}},
  year         = {2024},
  month        = dec,
  institution  = {Google DeepMind},
  note         = {Technical report on Gemini 2.X model family},
  url          = {https://storage.googleapis.com/deepmind-media/gemini/gemini_v2_5_report.pdf}
}

@misc{tyagi2024stepbystepreasoningsolvegrid,
      title={Step-by-Step Reasoning to Solve Grid Puzzles: Where do LLMs Falter?}, 
      author={Nemika Tyagi and Mihir Parmar and Mohith Kulkarni and Aswin RRV and Nisarg Patel and Mutsumi Nakamura and Arindam Mitra and Chitta Baral},
      year={2024},
      eprint={2407.14790},
      archivePrefix={arXiv},
      primaryClass={cs.CL},
      url={https://arxiv.org/abs/2407.14790}, 
}

@misc{golovneva2022roscoe,
      title={ROSCOE: A Suite of Metrics for Scoring Step-by-Step Reasoning}, 
      author={Olga Golovneva and Moya Chen and Spencer Poff and Martin Corredor and Luke Zettlemoyer and Maryam Fazel-Zarandi and Asli Celikyilmaz},
      year={2023},
      eprint={2212.07919},
      archivePrefix={arXiv},
      primaryClass={cs.CL},
      url={https://arxiv.org/abs/2212.07919}, 
}

@misc{prasad2023receval,
      title={ReCEval: Evaluating Reasoning Chains via Correctness and Informativeness}, 
      author={Archiki Prasad and Swarnadeep Saha and Xiang Zhou and Mohit Bansal},
      year={2023},
      eprint={2304.10703},
      archivePrefix={arXiv},
      primaryClass={cs.CL},
      url={https://arxiv.org/abs/2304.10703}, 
}

@misc{jacovi2024chainofthought,
      title={A Chain-of-Thought Is as Strong as Its Weakest Link: A Benchmark for Verifiers of Reasoning Chains}, 
      author={Alon Jacovi and Yonatan Bitton and Bernd Bohnet and Jonathan Herzig and Or Honovich and Michael Tseng and Michael Collins and Roee Aharoni and Mor Geva},
      year={2024},
      eprint={2402.00559},
      archivePrefix={arXiv},
      primaryClass={cs.CL},
      url={https://arxiv.org/abs/2402.00559}, 
}

@misc{lee2025evaluatingstepby,
      title={Evaluating Step-by-step Reasoning Traces: A Survey}, 
      author={Jinu Lee and Julia Hockenmaier},
      year={2025},
      eprint={2502.12289},
      archivePrefix={arXiv},
      primaryClass={cs.CL},
      url={https://arxiv.org/abs/2502.12289}, 
}

@misc{shi2023rethinkingicl,
      title={Rethinking the Role of Demonstrations: What Makes In-Context Learning Work?}, 
      author={Sewon Min and Xinxi Lyu and Ari Holtzman and Mikel Artetxe and Mike Lewis and Hannaneh Hajishirzi and Luke Zettlemoyer},
      year={2022},
      eprint={2202.12837},
      archivePrefix={arXiv},
      primaryClass={cs.CL},
      url={https://arxiv.org/abs/2202.12837}, 
}

@misc{li2024evaluatingmathematicalreasoninglarge,
      title={Evaluating Mathematical Reasoning of Large Language Models: A Focus on Error Identification and Correction}, 
      author={Xiaoyuan Li and Wenjie Wang and Moxin Li and Junrong Guo and Yang Zhang and Fuli Feng},
      year={2024},
      eprint={2406.00755},
      archivePrefix={arXiv},
      primaryClass={cs.CL},
      url={https://arxiv.org/abs/2406.00755}, 
}

@article{xue2023rcot,
  title={RCOT: Detecting and Rectifying Factual Inconsistency in Reasoning by Reversing Chain-of-Thought},
  author={Tianci Xue and Ziqi Wang and Zhenhailong Wang and Chi Han and Pengfei Yu and Heng Ji},
  journal={arXiv preprint arXiv:2305.11499},
  year={2023},
  url={https://arxiv.org/abs/2305.11499}
}

@article{zhao2023verifyandedit,
  title={Verify-and-Edit: A Knowledge-Enhanced Chain-of-Thought Framework},
  author={Ruochen Zhao and Xingxuan Li and Shafiq Joty and Chengwei Qin and Lidong Bing},
  journal={arXiv preprint arXiv:2305.03268},
  year={2023},
  url={https://arxiv.org/abs/2305.03268}
}

@misc{mao2025alympics,
      title={ALYMPICS: LLM Agents Meet Game Theory -- Exploring Strategic Decision-Making with AI Agents}, 
      author={Shaoguang Mao and Yuzhe Cai and Yan Xia and Wenshan Wu and Xun Wang and Fengyi Wang and Tao Ge and Furu Wei},
      year={2024},
      eprint={2311.03220},
      archivePrefix={arXiv},
      primaryClass={cs.CL},
      url={https://arxiv.org/abs/2311.03220}, 
}

@inproceedings{ltsz23inform,
title = "{INFORM} : Information e{N}tropy based multi-step reasoning {FOR} large language Models",
    author = "Zhou, Chuyue  and
      You, Wangjie  and
      Li, Juntao  and
      Ye, Jing  and
      Chen, Kehai  and
      Zhang, Min",
    editor = "Bouamor, Houda  and
      Pino, Juan  and
      Bali, Kalika",
    booktitle = "Proceedings of the 2023 Conference on Empirical Methods in Natural Language Processing",
    month = dec,
    year = "2023",
    address = "Singapore",
    publisher = "Association for Computational Linguistics",
    url = "https://aclanthology.org/2023.emnlp-main.216/",
    doi = "10.18653/v1/2023.emnlp-main.216",
    pages = "3565--3576",
}

@book{russell2020aima,
  title        = {Artificial Intelligence: A Modern Approach},
  author       = {Russell, Stuart and Norvig, Peter},
  year         = {2020},
  edition      = {4th},
  publisher    = {Pearson},
  address      = {Hoboken, NJ, USA},
  isbn         = {978-0134610993}
}

@misc{touvron2023llamaopenefficientfoundation,
      title={LLaMA: Open and Efficient Foundation Language Models}, 
      author={Hugo Touvron and Thibaut Lavril and Gautier Izacard and Xavier Martinet and Marie-Anne Lachaux and Timothée Lacroix and Baptiste Rozière and Naman Goyal and Eric Hambro and Faisal Azhar and Aurelien Rodriguez and Armand Joulin and Edouard Grave and Guillaume Lample},
      year={2023},
      eprint={2302.13971},
      archivePrefix={arXiv},
      primaryClass={cs.CL},
      url={https://arxiv.org/abs/2302.13971}, 
}

@misc{touvron2023llama2openfoundation,
      title={Llama 2: Open Foundation and Fine-Tuned Chat Models}, 
      author={Hugo Touvron and Louis Martin and Kevin Stone and Peter Albert and Amjad Almahairi and Yasmine Babaei and Nikolay Bashlykov and Soumya Batra and Prajjwal Bhargava and Shruti Bhosale and Dan Bikel and Lukas Blecher and Cristian Canton Ferrer and Moya Chen and Guillem Cucurull and David Esiobu and Jude Fernandes and Jeremy Fu and Wenyin Fu and Brian Fuller and Cynthia Gao and Vedanuj Goswami and Naman Goyal and Anthony Hartshorn and Saghar Hosseini and Rui Hou and Hakan Inan and Marcin Kardas and Viktor Kerkez and Madian Khabsa and Isabel Kloumann and Artem Korenev and Punit Singh Koura and Marie-Anne Lachaux and Thibaut Lavril and Jenya Lee and Diana Liskovich and Yinghai Lu and Yuning Mao and Xavier Martinet and Todor Mihaylov and Pushkar Mishra and Igor Molybog and Yixin Nie and Andrew Poulton and Jeremy Reizenstein and Rashi Rungta and Kalyan Saladi and Alan Schelten and Ruan Silva and Eric Michael Smith and Ranjan Subramanian and Xiaoqing Ellen Tan and Binh Tang and Ross Taylor and Adina Williams and Jian Xiang Kuan and Puxin Xu and Zheng Yan and Iliyan Zarov and Yuchen Zhang and Angela Fan and Melanie Kambadur and Sharan Narang and Aurelien Rodriguez and Robert Stojnic and Sergey Edunov and Thomas Scialom},
      year={2023},
      eprint={2307.09288},
      archivePrefix={arXiv},
      primaryClass={cs.CL},
      url={https://arxiv.org/abs/2307.09288}, 
}

\appendix
\section{LLM Input Prompt Example}
\label{sec:example_prompt}

The following example in Box~\ref{box:llm_prompt_example} illustrates the structured
prompt provided to the LLM during gameplay.
This prompt includes retrieved examples with corresponding optimal moves, the current
board state, and the list of available positions,
ensuring consistent input formatting and supporting in-context reasoning.

We note that numerous variations in prompt design and formatting are possible, and more
extensive prompt engineering could
potentially improve performance. In our experiments, we found that the simple coordinate
format $(x,y)$ worked reliably, both for
facilitating the LLM's output and for programmatically extracting moves from the textual
response.
An alternative would have been a structured JSON format with separate fields for $x$ and $y$,
but this was discarded after
initial testing as the simpler format provided sufficient accuracy
and consistency for our objectives.
Prompt optimization, while potentially beneficial, is outside the scope of this work.

\begin{tcolorbox}[
		colback=blue!4,
		colframe=blue!70!black,
		coltitle=white,
		fonttitle=\bfseries,
		title=LLM Prompt Example,
		label=box:llm_prompt_example,
		boxrule=1.1pt,
		arc=3mm,
		left=3.5mm, right=4mm, top=2mm, bottom=3mm
	]
	\begin{verbatim}
	
EXAMPLES OF BOARDS AND 
CORRISPONDING OPTIMAL MOVES:

Example 1: 
Board:
X | O | X
O | X | _
_ | _ | O
Optimal move: (2,2)

...

CURRENT GAME STATE: 
You are playing Tic-Tac-Toe as X.
The opponent is O.
Current board (3x3, _ for empty):
O | _ | X
_ | X | O
_ | _ | _

Empty positions (x,y): 
(0,1), (1,0), (2,0), (2,1), (2,2)

You are X.
Your next move in (x,y) format is:
\end{verbatim}

\end{tcolorbox}

\bigskip

\makeatletter
\@ifundefined{isChecklistMainFile}{
  \newif\ifreproStandalone
  \reproStandalonetrue
}{
  \newif\ifreproStandalone
  \reproStandalonefalse
}
\makeatother

\ifreproStandalone
\documentclass[letterpaper]{article}
\usepackage[submission]{aaai2026}
\setlength{\pdfpagewidth}{8.5in}
\setlength{\pdfpageheight}{11in}
\usepackage{times}
\usepackage{helvet}
\usepackage{courier}
\usepackage{xcolor}
\usepackage{comment}
\frenchspacing

\begin{document}
\fi
\setlength{\leftmargini}{20pt}
\makeatletter\def\@listi{\leftmargin\leftmargini \topsep .5em \parsep .5em \itemsep .5em}
\def\@listii{\leftmargin\leftmarginii \labelwidth\leftmarginii \advance\labelwidth-\labelsep \topsep .4em \parsep .4em \itemsep .4em}
\def\@listiii{\leftmargin\leftmarginiii \labelwidth\leftmarginiii \advance\labelwidth-\labelsep \topsep .4em \parsep .4em \itemsep .4em}\makeatother

\setcounter{secnumdepth}{0}
\renewcommand\thesubsection{\arabic{subsection}}
\renewcommand\labelenumi{\thesubsection.\arabic{enumi}}

\newcounter{checksubsection}
\newcounter{checkitem}[checksubsection]

\newcommand{\checksubsection}[1]{%
	\refstepcounter{checksubsection}%
	\paragraph{\arabic{checksubsection}. #1}%
	\setcounter{checkitem}{0}%
}

\newcommand{\checkitem}{%
	\refstepcounter{checkitem}%
	\item[\arabic{checksubsection}.\arabic{checkitem}.]%
}
\newcommand{\question}[2]{\normalcolor\checkitem #1 #2 \color{blue}}
\newcommand{\ifyespoints}[1]{\makebox[0pt][l]{\hspace{-15pt}\normalcolor #1}}

\section*{Reproducibility Checklist}

\vspace{1em}
\hrule
\vspace{1em}

\checksubsection{General Paper Structure}
\begin{itemize}

	\question{Includes a conceptual outline and/or pseudocode description of AI methods introduced}{(yes/partial/no/NA)}
	Yes, see Section~4 Method.

	\question{Clearly delineates statements that are opinions, hypothesis, and speculation from objective facts and results}{(yes/no)}
	Yes, see Section~6 Results and Discussion.

	\question{Provides well-marked pedagogical references for less-familiar readers to gain background necessary to replicate the paper}{(yes/no)}
	Yes, see Section~1 Introduction and Section~2 Related Work.

\end{itemize}
\checksubsection{Theoretical Contributions}
\begin{itemize}

	\question{Does this paper make theoretical contributions?}{(yes/no)}
	Yes, but the paper does not introduce a completly new formal theoretical method.
	The work relies on the theoretical assumption of entropy as a proxy for uncertainty, which
	is empirically validated in Section~6 Results and Discussion.

	\ifyespoints{\vspace{1.2em}If yes, please address the following points:}
	\begin{itemize}

		\question{All assumptions and restrictions are stated clearly and formally}{(yes/partial/no)}
		Yes, see Section~7 Limitations and Future Work.

		\question{All novel claims are stated formally (e.g., in theorem statements)}{(yes/partial/no)}
		NA.

		\question{Proofs of all novel claims are included}{(yes/partial/no)}
		NA.

		\question{Proof sketches or intuitions are given for complex and/or novel results}{(yes/partial/no)}
		NA.

		\question{Appropriate citations to theoretical tools used are given}{(yes/partial/no)}
		Yes, see Section~2 Related Work and Section~3 Method.

		\question{All theoretical claims are demonstrated empirically to hold}{(yes/partial/no/NA)}
		Yes, see Section~6 Results and Discussion where we empirically validate and discuss
		the use of entropy as a proxy for uncertainty.

		\question{All experimental code used to eliminate or disprove claims is included}{(yes/no/NA)}
		No, evaluation code will be publicly released upon publication.

	\end{itemize}
\end{itemize}

\checksubsection{Dataset Usage}
\begin{itemize}

	\question{Does this paper rely on one or more datasets?}{(yes/no)}
	No, we do not rely on any datasets.
	Optimal moves are precomputed using the Minimax algorithm for Tic-Tac-Toe.

	\ifyespoints{If yes, please address the following points:}
	\begin{itemize}

		\question{A motivation is given for why the experiments are conducted on the selected datasets}{(yes/partial/no/NA)}
		NA

		\question{All novel datasets introduced in this paper are included in a data appendix}{(yes/partial/no/NA)}
		NA

		\question{All novel datasets introduced in this paper will be made publicly available upon publication of the paper with a license that allows free usage for research purposes}{(yes/partial/no/NA)}
		NA

		\question{All datasets drawn from the existing literature (potentially including authors' own previously published work) are accompanied by appropriate citations}{(yes/no/NA)}
		NA

		\question{All datasets drawn from the existing literature (potentially including authors' own previously published work) are publicly available}{(yes/partial/no/NA)}
		NA

		\question{All datasets that are not publicly available are described in detail, with explanation why publicly available alternatives are not scientifically satisficing}{(yes/partial/no/NA)}
		NA

	\end{itemize}
\end{itemize}

\checksubsection{Computational Experiments}
\begin{itemize}

	\question{Does this paper include computational experiments?}{(yes/no)}
	Yes. For all subsequent questions in this subsection, if not otherwise specified, refer to Section~5 Experimental Setup.

	\ifyespoints{If yes, please address the following points:}
	\begin{itemize}

		\question{This paper states the number and range of values tried per (hyper-) parameter during development of the paper, along with the criterion used for selecting the final parameter setting}{(yes/partial/no/NA)}
		Yes, final hyperparameters were selected based on preliminary experiments.

		\question{Any code required for pre-processing data is included in the appendix}{(yes/partial/no)}
		NA

		\question{All source code required for conducting and analyzing the experiments is included in a code appendix}{(yes/partial/no)}
		No, evaluation code will be publicly released upon publication.

		\question{All source code required for conducting and analyzing the experiments will be made publicly available upon publication of the paper with a license that allows free usage for research purposes}{(yes/partial/no)}
		Partial, code will be publicly released upon publication.

		\question{All source code implementing new methods have comments detailing the implementation, with references to the paper where each step comes from}{(yes/partial/no)}
		Yes.

		\question{If an algorithm depends on randomness, then the method used for setting seeds is described in a way sufficient to allow replication of results}{(yes/partial/no/NA)}
		Yes.

		\question{This paper specifies the computing infrastructure used for running experiments (hardware and software), including GPU/CPU models; amount of memory; operating system; names and versions of relevant software libraries and frameworks}{(yes/partial/no)}
		Yes.

		\question{This paper formally describes evaluation metrics used and explains the motivation for choosing these metrics}{(yes/partial/no)}
		Yes.

		\question{This paper states the number of algorithm runs used to compute each reported result}{(yes/no)}
		Yes.

		\question{Analysis of experiments goes beyond single-dimensional summaries of performance (e.g., average; median) to include measures of variation, confidence, or other distributional information}{(yes/no)}
		Yes, see Section~6 Results and Discussion and Table~1.

		\question{The significance of any improvement or decrease in performance is judged using appropriate statistical tests (e.g., Wilcoxon signed-rank)}{(yes/partial/no)}
		Yes, see Section~6 Results and Discussion with Spearman and Kendall Tau correlation tests.

		\question{This paper lists all final (hyper-)parameters used for each model/algorithm in the paper’s experiments}{(yes/partial/no/NA)}
		Yes.

	\end{itemize}
\end{itemize}
\ifreproStandalone
\end{document}
\fi

\end{document}